\documentclass[conference]{IEEEtran}
\usepackage[numbers]{natbib}
\usepackage[bookmarks=true]{hyperref}
\usepackage{times}
\usepackage{multicol}
\usepackage{graphicx}
\usepackage{xcolor}
\usepackage{tabularx}
\usepackage{float}

\usepackage{amsmath,mathtools}
\usepackage{subcaption}
\usepackage{tikz}
\usetikzlibrary{scopes}
\graphicspath{{figures/}}

\title{Morphology dictates learnability in neural controllers}

\author{Joshua Powers, Ryan Grindle, Sam Kriegman, Lapo Frati, Nick Cheney, Josh Bongard\\University of Vermont}

\begin{document}
\maketitle
\begin{abstract}
Catastrophic forgetting continues to severely restrict the learnability of controllers suitable for multiple task environments. Efforts to combat catastrophic forgetting reported in the literature to date have focused on how control systems can be updated more rapidly, hastening their adjustment from good initial settings to new environments, or more circumspectly, suppressing their ability to overfit to any one environment. When using robots, the environment includes the robot's own body, its shape and material properties, and how its actuators and sensors are distributed along its mechanical structure. Here we demonstrate for the first time how one such design decision (sensor placement) can alter the landscape of the loss function itself, either expanding or shrinking the weight manifolds containing suitable controllers for each individual task, thus increasing or decreasing their probability of overlap across tasks, and thus reducing or inducing the potential for catastrophic forgetting.
\end{abstract}

\section{Introduction}
It has been shown in various single-task settings how an appropriate robot design can simplify the control problem \cite{lichtensteiger1999evolving,vaughan2004evolution,brown2010universal,bongard2011morphological,kriegman2019automated,PERVAN2019197}, but because these robots were restricted to a single training environment, they did not suffer catastrophic forgetting.

Catastrophic forgetting is a major and unsolved challenge in the machine learning literature \cite{french1999catastrophic,goodfellow2013empirical,kirkpatrick2017overcoming,masse2018alleviating}. Regardless of learning algorithm or task domain, a neural network trained to perform task A and then challenged with learning task B as well usually forgets A at the same rate as it learns B. Such interference can also occur when an agent attempts to learn tasks A and B simultaneously if gradients of improvement in A lead away from those of B.

In a multitask setting, \citet{powers2018effects} recently demonstrated that certain body plans suffer catastrophic forgetting, while others do not. It was hypothesized that a robot with the right morphology could in some cases alias separate tasks: certain designs are able to move in such a way that a seemingly different training instance converges sensorially to a familiar instance. However, this conjecture was not isolated and tested. Likewise, the relationship between the body and the loss landscape was not investigated.

In this paper, we provide a more thorough investigation on the role of embodiment in catastrophic forgetting based on the assumption that in order to avoid catastrophic forgetting, there must exist a set of control parameters that are adequately performant across multiple task environments simultaneously.
Since a robot's mechanical design can influence the set of controller parameters suitable for each individual task environment, we here test the hypothesis that a specific physical property of the robot's design|namely, the location of its sensors along its body|can help or hinder continual learning
by allowing for more or less overlap in suitable parameter settings across multiple task environments.

Using a simple yet embodied agent as our model, we analytically and empirically investigate how sensor location affects the weight manifolds of the neural controller over multiple tasks.
We show how morphological optimization often results in asymmetrical and unintuitive sensor arrangements with much more potential to allow learning algorithms to avoid catastrophic forgetting than more intuitive, symmetrical designs.
Thus, human designer bias, while often useful, can sometimes inadvertently increase the likelihood of catastrophic forgetting during learning.
This suggests that we should scrutinize our prior assumptions about the body plan of robots challenged with continual learning, and where possible replace them with end-to-end data-driven design automation.

\begin{figure*}
    {\sf \textbf{A} \hspace{16em} \textbf{B} \hspace{18em} \textbf{C}} \\
    {\centering
    \includegraphics[height=1.3in]{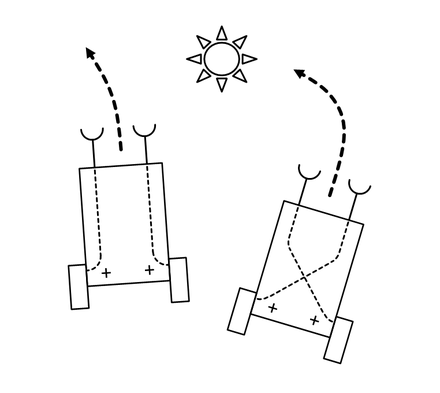}\hfill
    \includegraphics[height=1.3in]{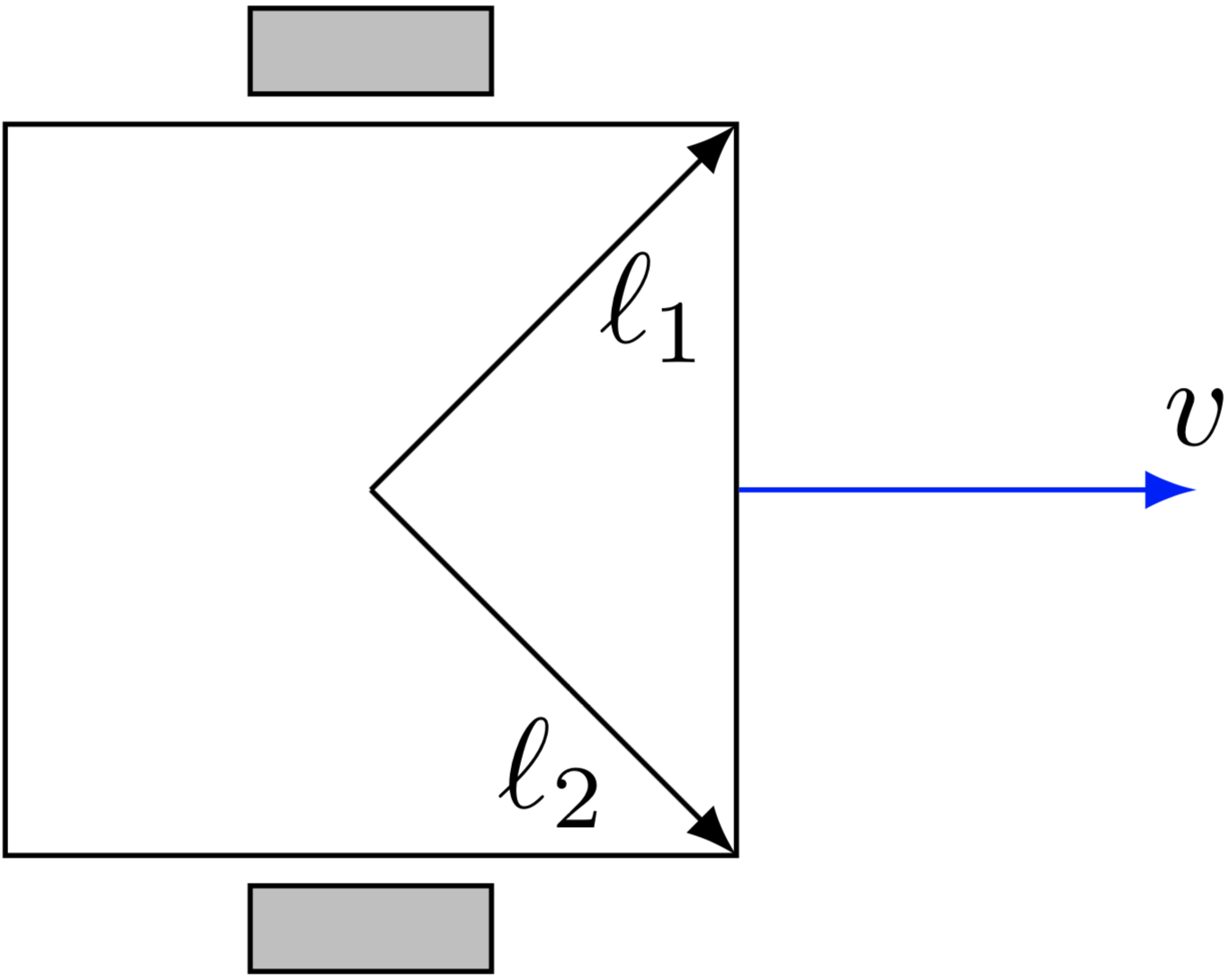}\hfill
    \includegraphics[height=1.3in]{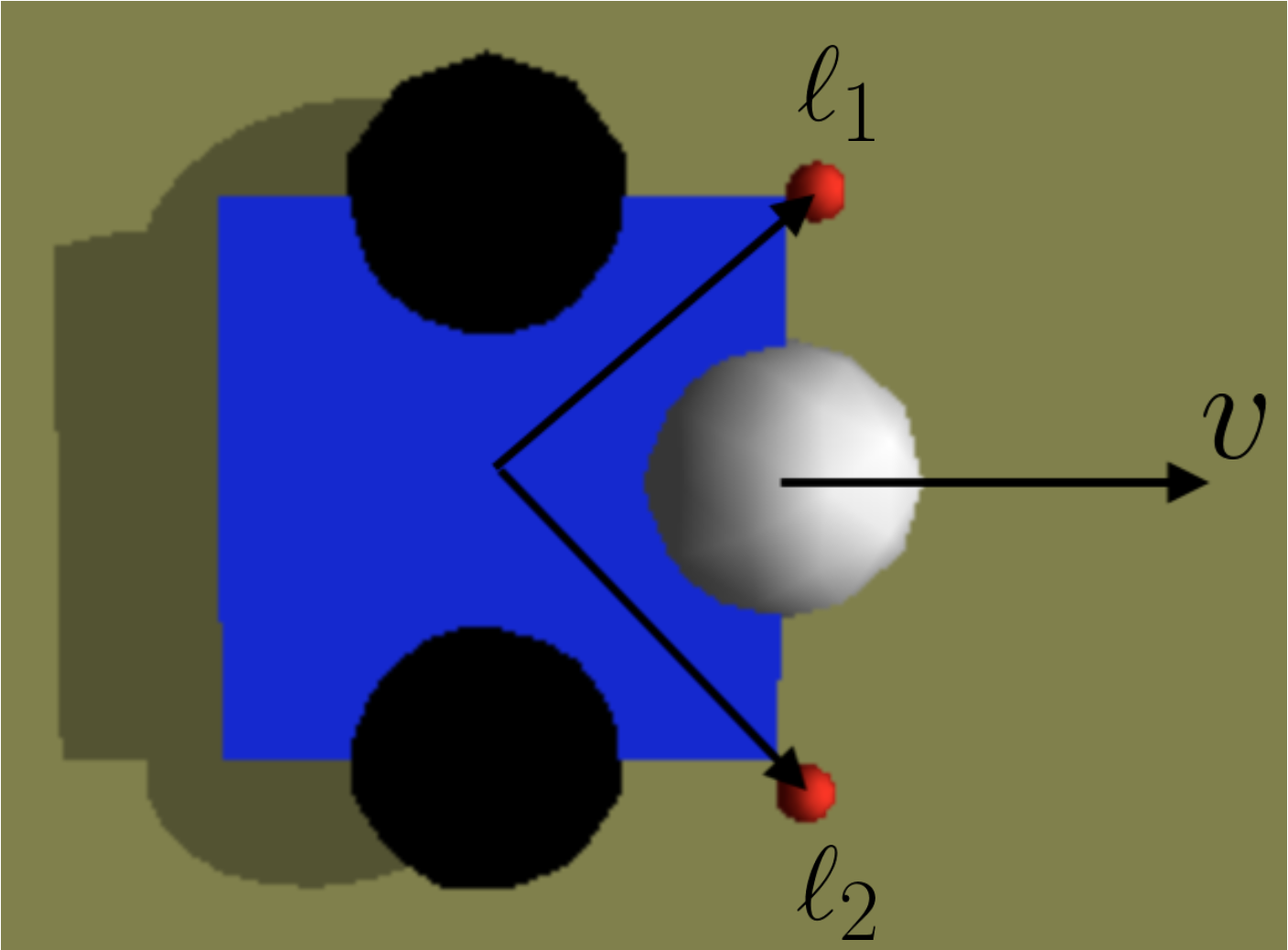}
    }
    \caption{\textbf{Modeling the robot.}
    \textbf{A:} The effect of lateral and contralateral synaptic connections (adopted from \cite{braitenberg1986vehicles}).
    \textbf{B:} The theoretical model with sensor positions determined by $\ell_1$ and $\ell_2$.
    \textbf{C:} The simulated robot with two light sensors (red), two motorized wheels (black), and a passive, anterior castor wheel for balance (gray).
    The robot is drawn (A-C) with symmetrical, anteriormost sensor placement, which we refer to in this paper as the ``canonical design''.
    }
    \label{fig:vehicles}
    
\end{figure*}
\section{Methods}
\subsection{The robot.}
\label{sec:vehicles}
The robot has a square frame, two separately-driven wheels, and two infrared sensors (Fig.~\ref{fig:vehicles}).
The sensors detect light according to the inverse square law, i.e., $1/d^2$, where $d$ is the distance from the light source;
occlusion was not modeled.
The motors driving the wheels are contralaterally connected to the sensors by weighted synapses yielding two trainable parameters $w_1, w_2 \in [-1.0, 1.0]$.

We here consider change to a single, isolated morphological attribute: the physical location of the two sensors, which can be placed anywhere on the dorsal surface of the robot's square body.
The location of the $i$-th sensor $\ell_i$ can be described by its Cartesian coordinates $\ell_i = (x, y)$, where $x, y \in [-0.5, 0.5]$, and (0, 0) denotes the center of the body (Fig.~\ref{fig:vehicles}B).

The effect of sensor location $\ell_i$ can be measured with respect to the space, denoted $\theta$, of possible synapse weight pairs ($w_1$, $w_2$).
Since we cannot perform an exhaustive sweep over the infinitude of possible sensor positions,
we discretized each dimension of $\ell_i$ into nine uniformly-spaced bins.
Because sensors are varied in two dimensions ($x$ and $y$) there are $9^2=81$ possible locations for each sensor;
and because there are two such sensors, the space $\theta$ is discretized into a 81-by-81 uniformly-spaced grid, thus yielding a searchable space of 6561 possible robot designs.

For each of the 6561 designs, we conducted another sweep over the synapse weights $(w_1, w_2)$, likewise discretizing each weight into 121 evenly-space values, yielding $121^2 = 14641$ different weight configurations.
Finally, for each of the $6561\times14641=96059601$ evaluated combinations of sensor locations and weight values, we analyzed the robot analytically using differential equations and empirically using a physics engine. These discretizations were chosen to be as small as possible within the limit of our computational resources and time.


\subsection{The task environments.}
\label{sec:environments}
The task is phototaxis in four environments, which differ in their position of the light source in relation to the robot. 
The light source is placed at polar coordinates $(r, \varphi)$ where $\varphi \in \{45^{\circ}, 135^{\circ}, 225^{\circ}, 315^{\circ}\}$ and $r$ is a fixed distance. 
A controller was considered successful for a given environment if the robot comes within 0.2~cm of the light source at any time during its evaluation period.

While there is of course a general strategy that solves the task for all environments (follow the light), the easiest gradients to follow in the loss landscape are initially those which produce forward locomotion in a single direction and cause the robot to ignore the light. 
This is because, from the robot's perspective, due to the inverse square law of light decay, improving its ability to move in the one environment with least loss earns quadratically
more reward than improvements to locomotion in any of the other three environments in which the robot is less proficient. This causes the catastrophic forgetting experienced by neural learning algorithms.

\subsection{The metrics.}
\label{sec:metrics}
We here define two metrics: $M_L$ and $M_{CF}$, that are measured over $k=4$ environments. These metrics measure how a robot design impacts the weight space of the controller and consequently measure how amenable to learning a robot would have been if the controller were to be learned with a standard learning algorithm rather than found by grid search.
$M_L$ measures controller learnability: how easy it would be to learn a generalist controller. 
$M_{CF}$ measures resistance to catastrophic forgetting: the probability that a environment-specific controller will generalize to other environment.

For each mechanical design $(\ell_1, \ell_2)$, we expect some optimal manifolds $\theta_k^*$ in the space of control parameters $(w_1, w_2)$ to succeed for a specific environment $k$. For a controller to be successful in multiple environments, it must reside within the intersection of environment-specific manifolds, $theta^*$, on the loss surface. Thus, the likelihood of finding a generalist controller|its learnability|will be proportional to the size of the intersection ($M_L$). Likewise, a controller's potential to resist catastrophic forgetting ($M_{CF}$) will be proportional to the ratio of generalist controllers (those successful in all four environments) to specialists (those successful in at least one environment).

\begin{figure}[t]
    \centering
    \includegraphics[width=0.8\linewidth]{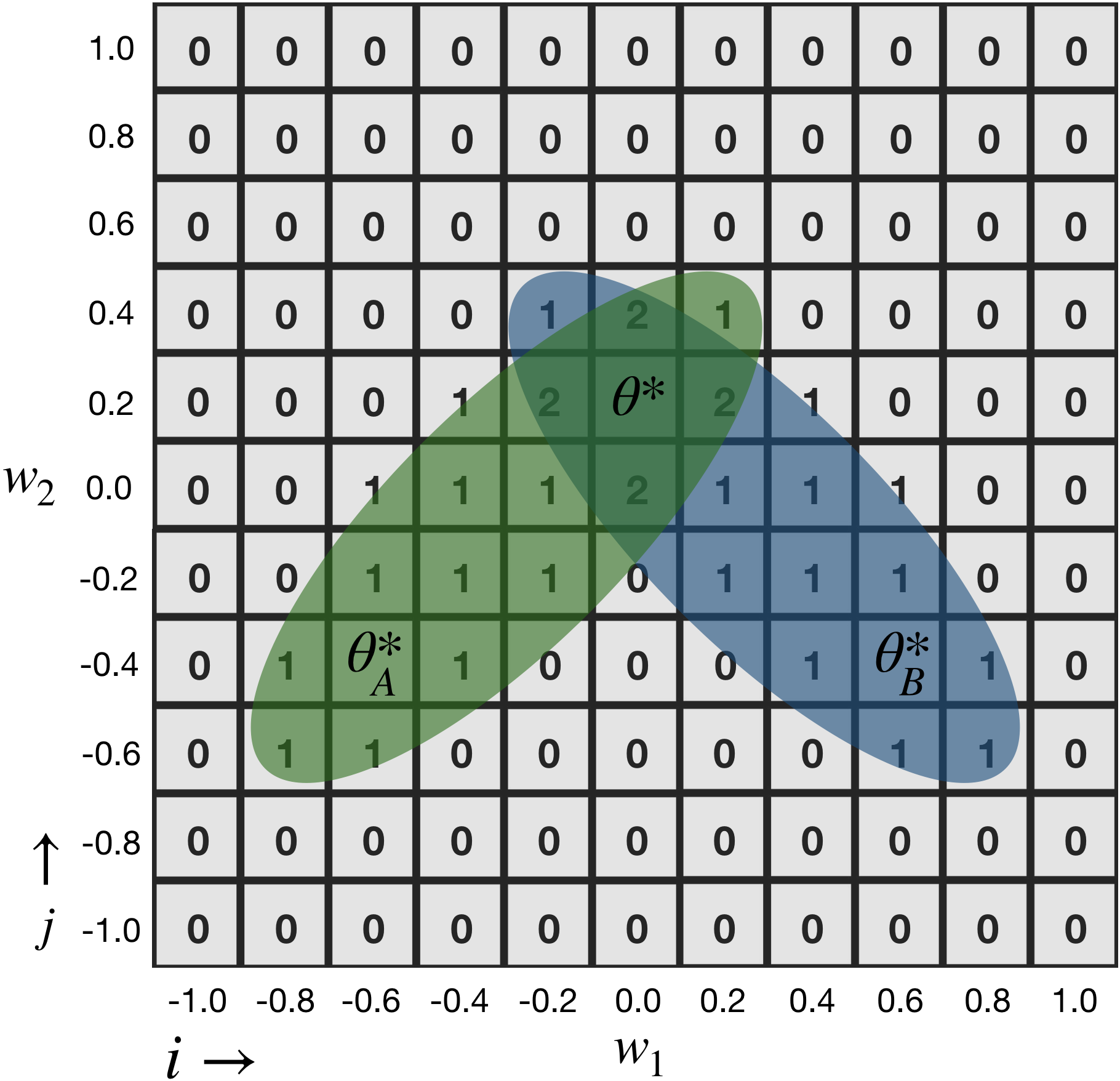}
    \caption{A general example of overlapped binary success matrices for some tasks A and B.
    Each element represents a different controller. Generalist controllers sit inside the intersection $\theta^*$ of successful environment-specific controllers $\theta_k^*$.}
    \label{fig:metrics}
\end{figure}

Given a design $(\ell_1, \ell_2)$ and environment $k$, a binary success matrix $S^{k}(\ell_1, \ell_2)$ is constructed such that each element $S^{k}_{i, j}(\ell_1, \ell_2)$ is either 1 (success) or 0 (failure). By overlapping the success matrices for a fixed design across the four environments, we can visualize the manifolds $\theta_k^*$ where $k\in\{1, 2, 3, 4\}$ for the robot (Fig.~\ref{fig:metrics}).

We define the overlap $\mathcal{O}$ as a element-wise sum of the success matrices over each environment $k$:
\begin{equation}
    \label{eq:overlap}
    \mathcal{O} = \sum_{k=1}^4 S^k(\ell_1, \ell_2) .
\end{equation}

The learnability metric is simply the proportion of 4s (where a 4 represents success in all 4 environments) in the overlapped success matrices to the entire matrix space:
\begin{equation}
    \label{eq:metric1}
    M_L = \frac{g_4(\mathcal{O})}{n^2},
\end{equation}
where $g_k$ is a function that counts the total elements of a matrix with value equal to $k$ and $n$ is the square dimension of the matrix defined by the discrete parameter sweep.

Resistance to catastrophic forgetting is measured by:
\begin{equation}
    \label{eq:metric2}
    M_{CF} = \begin{cases}
    0 \; \text{ if } \mathcal{O} \text{ is a null matrix}, \\
    g_4(\mathcal{O}) \left[\sum_{k=1}^4 g_k(\mathcal{O})\right]^{-1}
    \; \text{otherwise}.
    \end{cases}
\end{equation}
which is the number of control parameters that solved all four environments divided by the number of control parameters that solved at least one.

\subsection{The theoretical model.}
\label{sec:theoretical}
The location and orientation of the robot can be defined by a system of differential equations, where the change in position and orientation is determined by the change in light captured by two sensors. 
Ignoring deviations from the idealized environment, such as sensor noise and friction,
the rate of angular and linear velocities will be proportional to a linear combination of the sensor values.

Let $\alpha(t)$ be the angle of the robot at time $t$, where $\alpha=0$ denotes the positive $x$ direction, and $\phi(t) = (x(t), y(t))$ be the position of the robot in the world, then if the robot is located at the origin and facing east ($\alpha = 0$), its two light sensors are located exactly at $\ell_1$ and $\ell_2$, and they each capture some amount of light $s_1(t)$ and $s_2(t)$, respectively. 

Hence the absolute position of the $i$-th sensor is $\phi(t) + R_\alpha \ell_i^{\hspace{3pt}T}$, where
	\begin{equation}
		R_\alpha = \begin{bmatrix}
			\cos \alpha & -\sin \alpha \\
			\sin \alpha & \cos \alpha \\
		\end{bmatrix}
	\end{equation}
is the two-dimensional counterclockwise rotation matrix (in the amount $\alpha$). 

If we formulate the problem such that it is the robot's initial position and heading that is adjusted in each environment, instead of the position of the light source, we can assume that the source is always at the origin. Then, the distance of $\ell_i$ from the light source is given by:  \mbox{$\|\phi(t) + R_\alpha \ell_i^{\hspace{3pt} T}\|$.}
And since the intensity of light is inversely proportional to the square of the distance, the sensor values are given by: 
\begin{equation}
\label{eq:sensor_value}
s_i(t) = c \cdot \|\phi(t)^T + R_\alpha {\ell_i}^T\|^{-2},
\end{equation}
where $c$ is a constant that we set equal to one. 

Assuming the robot turns based on the difference between the sensor values (with weights applied), the velocity of the robot is the average of the two sensor values. 
Thus, the following system of equations determines the location and orientation of the robot:
	\begin{equation}
	    \label{eq:ode}
		\begin{cases}
			\dot{x} = v(t) \cos \alpha	\\
			\dot{y} = v(t) \sin \alpha \\
			\dot{\alpha} = w_1s_1(t) - w_2s_2(t), \\
		\end{cases}
	\end{equation}
where $v$ is the velocity of the robot given by $2v(t) = w_1s_1(t)+w_2s_2(t).$

\subsection{The empirical model.}
\label{sec:empirical}
Because our theoretical model is highly abstracted from the real world and built on a number of assumptions (no friction, motor limits, collisions, etc.) which may potentially affect the robot's behavior, we also empirically test our claims by simulating the robots inside a physics engine.

The robot is simulated using Open Dynamics Engine
(Fig.~\ref{fig:vehicles}C).
Just like the theoretical model, the simulated robot contains two light sensors, which innervate two motorized, spherical wheels (each with a single axis of rotation), which are attached midway along the sides of a $1 \times 1 \times 0.13$~cm box.
Additionally, an anterior passive castor wheel was added for balance. Finally, a light source is simulated on the floor of the environment at polar coordinates $(r, \alpha)$ as a fixed sphere with radius 0.2 cm. In simulation, the behavior of a robot in a given environment is taken to be successful if it collides with the light source at anytime during an evaluation period of 2500 time steps ($dt=0.05$) or 125 seconds.

In order to replicate the baseline behavior of the canonical robot design it was necessary to pre-optimize various physical attributes of the robot's body, including the mass of each component, the radii of the wheels, and the maximum torque, speed, and target actuation rate. A multiobjective optimization algorithm \cite{hadka12b} was used to find a base morphology, with the sensors fixed in the canonical position, that was both performant and stable. The first objective was to maximize the performance of the robot (distance from the light source), summed across all the four environments. The second objective was to minimize the sum of the maximum torque, speed and target actuation rate. This second objective is used to avoid both simulator instability and behavior that is unlikely to transfer to reality.

After discovering a good base morphology, we performed the nested grid search described in \S\ref{sec:vehicles}, for sensor locations ($\ell_1$, $\ell_2$) and weights ($w_1, w_2$). 

\begin{figure*}[t]
    {\sf \textbf{A} \hspace{17em} \textbf{B} \hspace{19em} \textbf{C}}\\
    {\centering
    \includegraphics[height=1.3in]{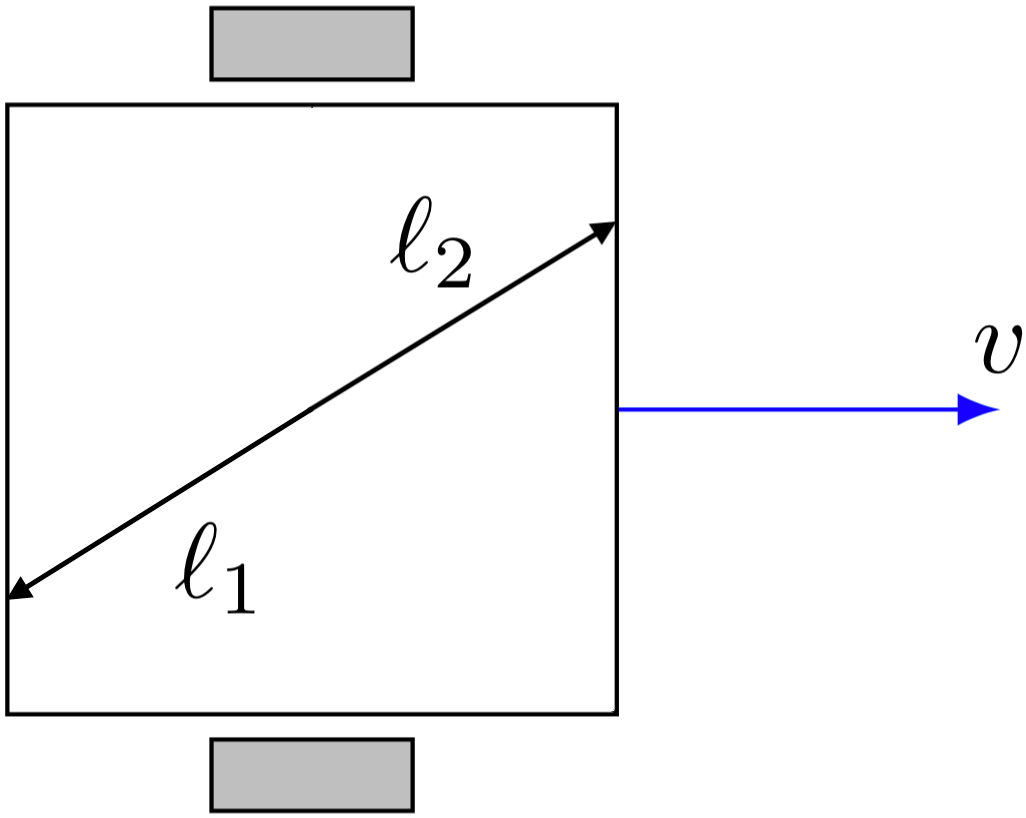}\hfill
    \includegraphics[height=1.3in]{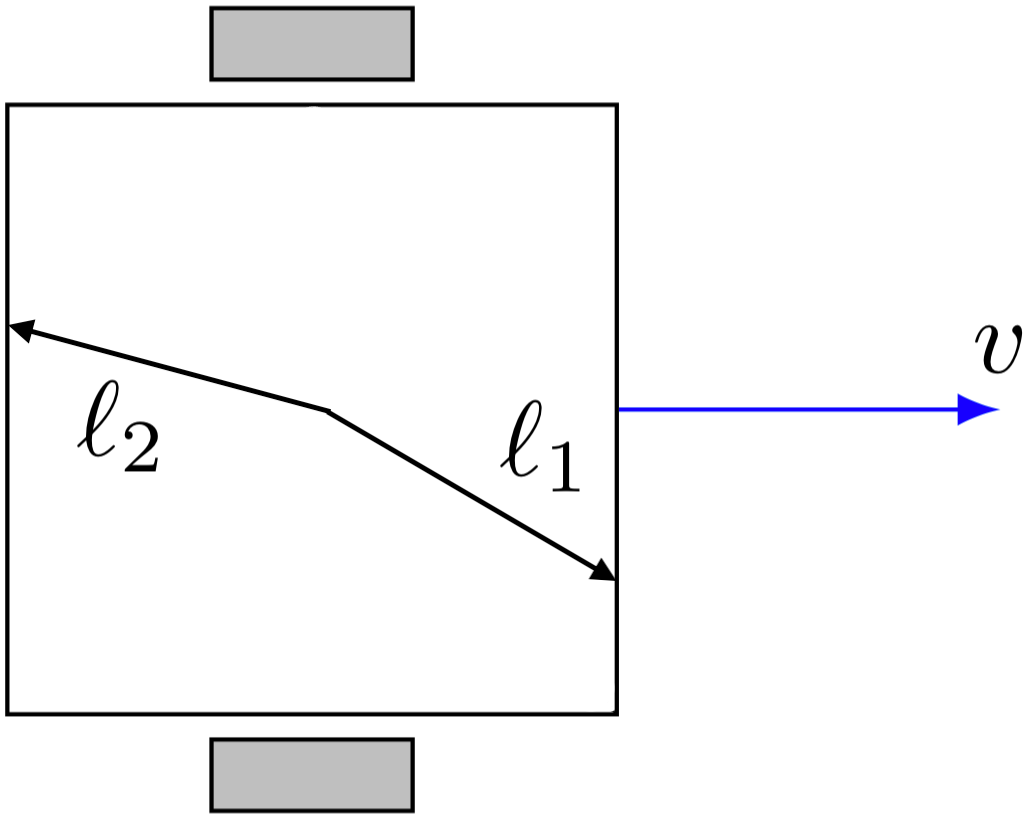}\hfill
    \includegraphics[height=1.3in]{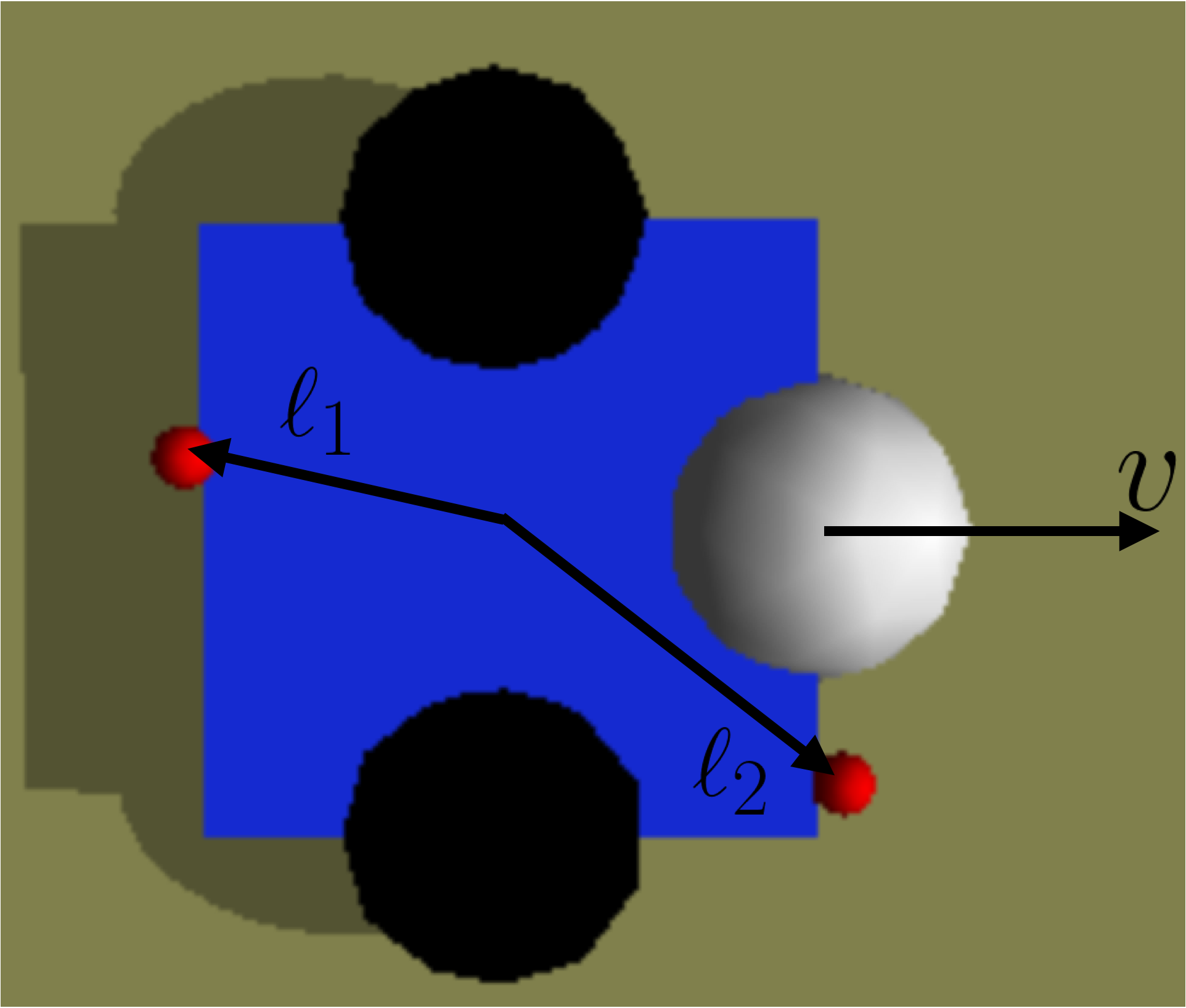}
    }
    \caption{\textbf{The best designs} under the 
    theoretical model according to controller learnability (\textbf{A}; Metric $M_L$) and resistance to catastrophic forgetting (\textbf{B}; Metric $M_{CF}$).
    Under the empirical model, the design with the highest controller learnability was also the most resistant to catastrophic forgetting (\textbf{C}).
    Although the design space we swept over contains many symmetrical sensor arrangements, and most real robots utilize symmetrical sensor distributions,
    the best designs are notably asymmetrical.
    }
    \label{fig:best_morphology}
\end{figure*}
\section{Results}

\subsection{Theoretical results.}
We employed SciPy (\textit{scipy.integrate.odeint}) for numerical integration of the robot's location and orientation (Eq.~\ref{eq:ode}), for $10^5$ timesteps.

For each evaluated mechanical design and controller (sensor locations and synapse weights), the robot's trajectory is computed in each of the four environments defined in \S\ref{sec:environments}. 
As in the empirical model, if robot's trajectory comes within 0.2 units of the light source, the robot is determined to have succeeded in that environment. 
Otherwise, it is determined to have failed.

The mechanical design sketched in Fig.~\ref{fig:best_morphology}A (and its mirror image when reflected about the sagittal plane) had the highest controller learnability score, with $M_L = 0.286$. However it did score the best in resistant to catastrophic forgetting: the proportion of resistant to nonresistant controllers for that design was $M_{CF}=0.636$, whereas several other found designs had full resistance $M_{CF}=1$. But those with a perfect ratio $M_{CF}=1$ had much smaller optimal weight manifold: the highest learnability score achieved by this group was $M_L=0.206$. 
In other words, while all the successful environment-specific controllers for these designs generalize across all four environments, the manifold containing them is much smaller and thus would be more difficult to find if controller parameters were to be optimized by learning.

The canonical design had a much lower controller learnability ($M_L=0.049$) and resistance to catastrophic forgetting ($M_{CF}=0.24$), than many found asymmetrical designs.

\begin{figure}[t]
    \centering
    \includegraphics[width=\linewidth]{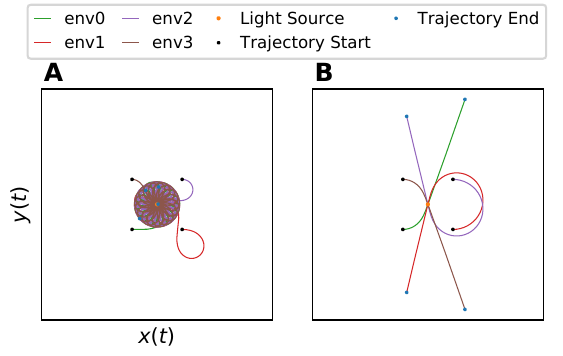}
    \caption{\textbf{Successful trajectories with canonical (symmetrical) sensor location under the theoretical model.} 
    With canonical sensor placement $ \{ \ell_1 = (0.5, 0.5)$, $\ell_2 = (0.5, -0.5) \} $ (Fig.~\ref{fig:vehicles}B),
    only 57 of the $121^2$ evaluated controllers (0.4\%) were successful all four environments.
    \textbf{A:} The trajectories generated by one of the successful controllers $(w_1, w_2) = (0.6, 0.98)$.
    This controller initially generated phototaxis, but passed through the light source and continued to move away from it.
    \textbf{B:} The trajectories generated by another successful controller $(w_1, w_2) = (0.77, 0.77)$. 
    This controller continuously spirals about the light source.
    The light source is drawn once at the origin, and the initial positions/orientations of the robot relative to the it are superimposed for the four environments.
    }
    \label{fig:canonical_trajectories}
\end{figure}

\begin{figure}[t]
    \centering
    \includegraphics[width=\linewidth]{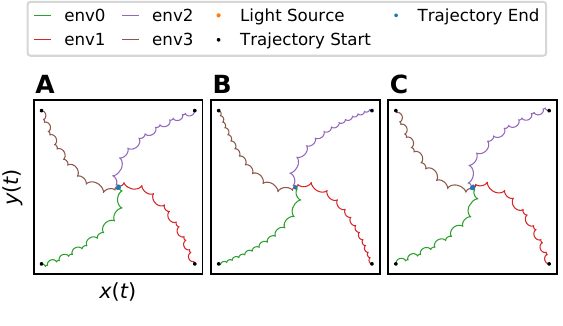}
    \caption{\textbf{Trajectories of the design with maximal controller learnability, as measured by metric $M_1$ (Eq.~\ref{eq:metric1}) under the theoretical model.} 
    With sensor locations $\ell_1 = (-0.5, -0.25)$ and $\ell_2 = (0.5, 0.25)$ (Fig.~\ref{fig:best_morphology}A), 2255 of the $121^2$ evaluated controllers (15.4\%) were successful all four environments. 
    \textbf{A:} The trajectories generated by one of the successful controllers, parameterized by weights $(w_1, w_2) = (-0.85, 0.82)$.
    \textbf{B:} The trajectories with weights $(-0.8, 0.6)$.
    \textbf{C:} The trajectories with weights $(-0.28, 0.37)$. 
    The axes are equivalent to those in Fig.~\ref{fig:canonical_trajectories}.
    }
    \label{fig:best_trajectories}
\end{figure}

For both the canonical, symmetrical design (Fig.~\ref{fig:canonical_trajectories}) and the design with the highest controller learnability score (Fig.~\ref{fig:best_trajectories}) there are initial conditions that generate persistent phototaxis: the robot moves toward the light source and remains near it.
However, whereas 35 of the 35 found phototaxing controllers for the found design remain in the neighborhood of the source, only 2 of the 6 found controllers for the canonical design do so.
Some initial conditions of the canonical design initially produce phototaxis, but the design passes through the source and then continues to move away from it (Fig.~\ref{fig:canonical_trajectories}A). This was not observed to occur with the ``optimized'' designs.

\subsection{Empirical results.}
As with the theoretical model the empirical model showed that non-intuitive asymmetrical designs scored higher in learnability and in resistance to catastrophic forgetting. However unlike the theoretical model one design performed the best on both metrics.

The found asymmetrical design shown in Fig.~\ref{fig:best_morphology}C  
had both the highest generalist controller learnability ($M_L = 0.0039$) and resistance to catastrophic forgetting ($M_{CF} = 0.038$). 
Overall, there were 57 generalist phototaxing controllers found (out of 14641 evaluated; 0.389\%) for this design,
compared to only one generalist phototaxing controller found (0.0068\%) for the canonical, symmetrical design.
The controller learnability of the canonical design was thus $M_L=0.000068$; and its resistance to catastrophic forgetting was $M_{CF}=0.00052$.
Thus, the found asymmetrical design has both higher controller learnability and resistance to catastrophic forgetting.

\subsection{Overview.}
\begin{figure*}
    \centering
    \includegraphics[width=0.93\linewidth]{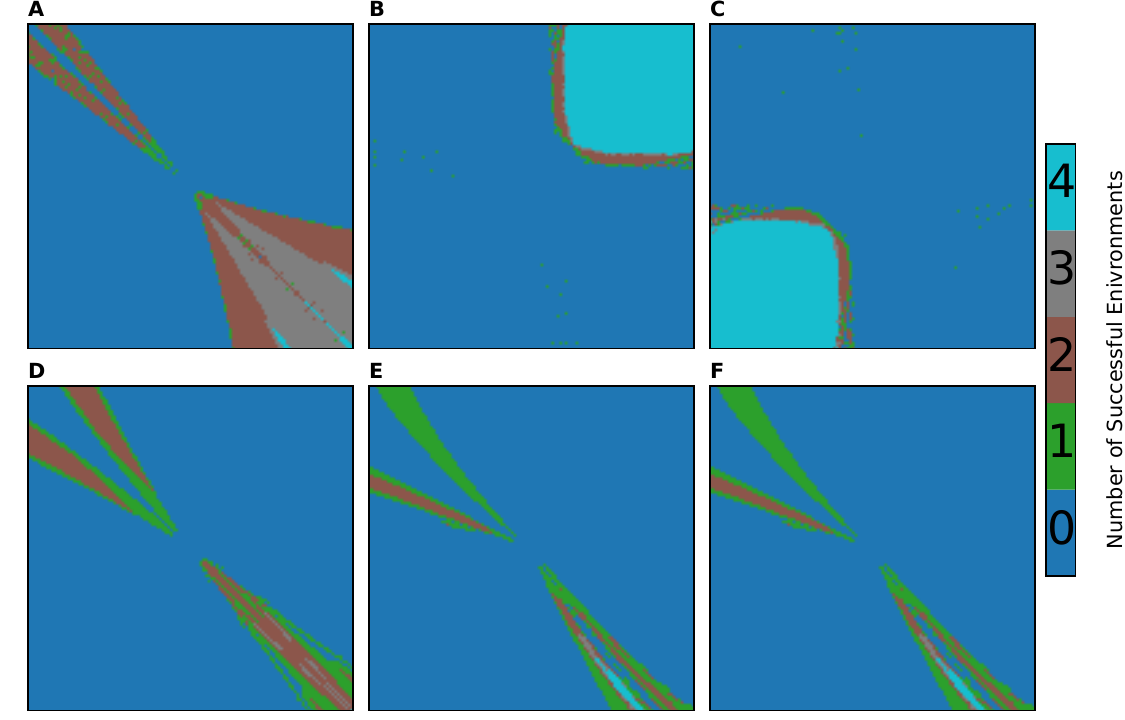}
    \caption{\textbf{Measuring learnability and forgetting.}
    For both the theoretical (\textbf{A-C}) and empirical models (\textbf{D-F}), 
    we performed a 121-by-121 grid search of controller weights (14641 unique controllers) nested within a 81-by-81 grid search for sensor locations (6561 unique designs).
    The controller space its mapped for the canonical, symmetrical design (A, D), the design with highest controller learnability ($M_L$; Eq.~\ref{eq:metric1}) (B, E), and the design most resistant to catastrophic forgetting ($M_{CF}$; Eq.~\ref{eq:metric2}) (C, F).
    Under the controller sweep on (D-F), the design with the highest controller learnability also had the greatest resistance to forgetting, so E and  are identical.
    Each pixel represents a different controller $(w_1, w_2)$ for the given design, and is colored by the number of environments that the combination successfully exhibited phototaxis 
    (i.e., the overlapped binary success matrices, defined by Eq.~\ref{eq:overlap}).
    Under both the theoretical and empirical models,
    the unintuitive asymmetrical designs (B, E) were found to have higher controller learnability and greater resistance to forgetting in their landscape than their respective canonical design (A, D) as measured by the number pixels in the heatmap that are successful in all four environments (cyan).
    Likewise, the asymmetrical designs (C, F) had higher resistance to catastrophic forgetting as measured by the number of cyan pixels to non-blue pixels.
    }
    \label{fig:landscape}
\end{figure*}

In Fig.~\ref{fig:landscape} the successes of weight manifolds for all of these design in both the theoretical and empirical model can be seen in detail, where cyan represents weight assignments that succeed in all for environments for a given design. These weight manifolds show clearly that in this case the weight assignments for the asymmetrical would be much easier to find by a learning algorithm while the canonical design is akin to looking for a needle in a haystack.

\begin{figure*}
    \centering
    \includegraphics[width=\textwidth]{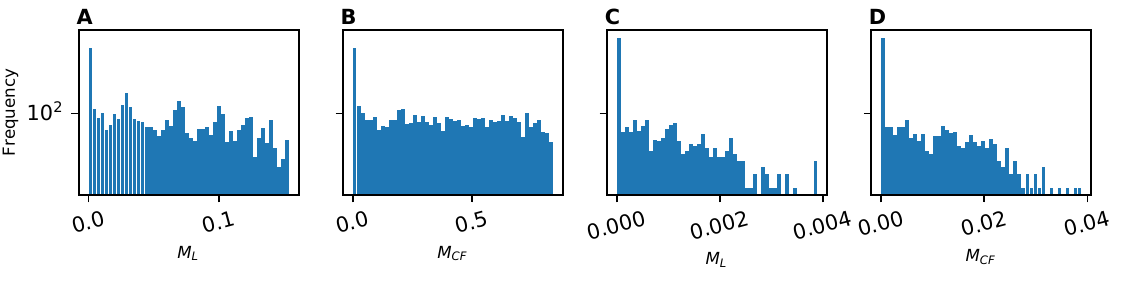}
    \vspace{-2em}
    \caption{\textbf{Measuring successful multitask learning.}
    The distribution of metrics $M_L$ and $M_{CF}$, for all evaluated designs, in the theoretical (A, B), empirical (C, D).
    Metric $M_L$ (Eq.~\ref{eq:metric1}) indicates controller learnability: the proportion of controllers deemed successful in all four environments, for a given design.
    Metric $M_{CF}$ (Eq.~\ref{eq:metric2}) indicates resistance to catastrophic forgetting: the ratio of the number of controllers successful in all environments, over the number successful in at least one. If no controllers are successful for a given design, $M_{CF}=0$.}
    \label{fig:histogram}
\end{figure*}

Fig.~\ref{fig:histogram} plots the frequency of metrics $M_L$ and $M_{CF}$ (Eqs.~\ref{eq:metric1} and \ref{eq:metric2}, respectively) within each bin of the grid search. This again shows how there are many designs (including intuitive symmetric ones) that score poorly on $M_L$ and $M_{CF}$ while there are relatively few designs that perform well. Thus a given design has a drastic effect on the theoretical learnability of a robots controller parameters.

\vspace{-1.0em}
\section{Discussion}
In this paper, we considered a simple robot and task in order to sample the entire loss landscape of the weight manifold at a relatively high resolution. 
While we haven't tested these robot with any specific learning algorithm, our results suggest that changes in one element of a robot's design (sensor location) can fundamentally alter the loss surface, thus influencing the controller's learnability, and resistance to catastrophic forgetting. 
More specifically, by changing sensor location, we observed changes in the number and placement along the loss surface of control parameters suitable for individual environments, as well in how these optimal yet environment-specific parameters overlapped across different environments to produce generalist controllers which resist catastrophic forgetting. However, we acknowledge that this work mainly builds a theoretical foundation and that our metrics need to be tested against existing methods for learning.

Previous efforts to avoid catastrophic forgetting have relied almost exclusively on increased control complexity. 
Most were focused on making changes to small subsets of neural network weights \cite{kirkpatrick2017overcoming,masse2018alleviating,french1991using,robins1995catastrophic,he2018overcoming,beaulieu2018combating,schwarz2018progress,titsias2019functional}. 
Others have attempted to sidestep the problem by learning good initial weights such that they can be quickly updated when switching between tasks \cite{finn2017model,gidaris2018dynamic}. 
We have shown here that, in theory, regardless of the algorithm used it is also possible to alleviate catastrophic forgetting by changing aspects of the robot's design, without increasing control complexity, but doing so can be non-intuitive.

We found that even the seemingly trivial case of phototaxis with contralateral connections described by \citet{braitenberg1986vehicles} can require morphological tuning to work as expected in a single simulated environment, and that, when challenged to perform in additional environments, other adjustments in morphology, specifically to sensor location, could either suppress or multiply the potential for catastrophic forgetting by expanding or shrinking the overlap of performant controller settings for that body plan across different task environments.

The physical location of sensors is thus a relevant property of robots that is nevertheless abstracted away in the (mostly disembodied) systems that address catastrophic forgetting reported in the literature to date.
While sensor location could in principle be dynamically controlled via a lattice of sensors \cite{kramer2011wearable} or adjustable antenna \cite{fend2003active}, 
change in (and rational control over) other morphological attributes|such as geometry \cite{kriegman2019automated}, material properties \cite{narang2018transforming}, or the number and placement of actuators \cite{lipson2000automatic}|is much more difficult in practice, and such design elements are almost always presupposed and fixed prior to training \cite{cheney2018scalable}.
\\ \indent
However, unless experimental proof is obtained in the real world, this theory will remain speculation.
In fact it is possible that the proposed empirical model using rigid body physics was more disconnected from reality than our theoretical model.
The simulated wheels, for instance, have just a single point of contact with the ground.
A more realistic surface contact geometry might completely change the optimal sensor locations, 
but there's also reason to believe that the loss surface manifolds containing adequate controllers for a compliant body could be larger than those of a rigid body \cite{kriegman2019automated,hauser2011towards}, further increasing the probability of overlap across tasks.
\\ \indent
In the limit, machines with the right morphology may use a single controller to accomplish a set of tasks that appear 
disparate to a robot with a different body plan.
For example, a granular jamming gripper \cite{brown2010universal} need not precisely control the placement of each joint around differently shaped objects: a single policy (vacuum air, hold, relax) works regardless of object shape.
However, this control policy is exceedingly simple.
The degree to which morphology influences learnability in more complex robots,
task environments and behaviors has yet to be investigated, but will be the focus
of future work.
\\ \indent
In this work, two control- and two morphology parameters 
were optimized. In future work we will investigate whether 
co-optimizing the morphology and control parameters
confers greater overall learnability on the robot compared to a robot with a fixed
morphology and four control parameters. This will help determine whether a poorly
chosen mechanical design can be compensated for by increased control complexity.

\section*{Acknowledgements}
This work was supported by NSF EFRI award 1830870, and by the Vermont Space Grant Consortium under NASA Cooperative Agreement NNX15AP86H. Computational Resources were provided by The Vermont Advanced Computing Core (VACC). Thanks is also due to the beautiful wife of the first author whose continual support makes work like this possible.
\bibliographystyle{plainnat}
\bibliography{main.bib}

\begin{thebibliography}{27}
\providecommand{\natexlab}[1]{#1}
\providecommand{\url}[1]{\texttt{#1}}
\expandafter\ifx\csname urlstyle\endcsname\relax
  \providecommand{\doi}[1]{doi: #1}\else
  \providecommand{\doi}{doi: \begingroup \urlstyle{rm}\Url}\fi

\bibitem[Beaulieu et~al.(2018)Beaulieu, Kriegman, and
  Bongard]{beaulieu2018combating}
Shawn~LE Beaulieu, Sam Kriegman, and Josh~C Bongard.
\newblock Combating catastrophic forgetting with developmental compression.
\newblock In \emph{Proceedings of the Genetic and Evolutionary Computation
  Conference}, pages 386--393. ACM, 2018.
\newblock URL \url{https://dl.acm.org/citation.cfm?id=3205455.3205615}.

\bibitem[Bongard(2011)]{bongard2011morphological}
Josh Bongard.
\newblock Morphological change in machines accelerates the evolution of robust
  behavior.
\newblock \emph{Proceedings of the National Academy of Sciences}, 108\penalty0
  (4):\penalty0 1234--1239, 2011.
\newblock URL \url{https://www.pnas.org/content/108/4/1234}.

\bibitem[Braitenberg(1986)]{braitenberg1986vehicles}
Valentino Braitenberg.
\newblock \emph{Vehicles: Experiments in synthetic psychology}.
\newblock MIT press, 1986.
\newblock URL \url{https://mitpress.mit.edu/books/vehicles}.

\bibitem[Brown et~al.(2010)Brown, Rodenberg, Amend, Mozeika, Steltz, Zakin,
  Lipson, and Jaeger]{brown2010universal}
Eric Brown, Nicholas Rodenberg, John Amend, Annan Mozeika, Erik Steltz,
  Mitchell~R Zakin, Hod Lipson, and Heinrich~M Jaeger.
\newblock Universal robotic gripper based on the jamming of granular material.
\newblock \emph{Proceedings of the National Academy of Sciences}, 107\penalty0
  (44):\penalty0 18809--18814, 2010.
\newblock URL \url{https://www.pnas.org/content/107/44/18809}.

\bibitem[Cheney et~al.(2018)Cheney, Bongard, SunSpiral, and
  Lipson]{cheney2018scalable}
Nick Cheney, Josh Bongard, Vytas SunSpiral, and Hod Lipson.
\newblock Scalable co-optimization of morphology and control in embodied
  machines.
\newblock \emph{Journal of The Royal Society Interface}, 15\penalty0
  (143):\penalty0 20170937, 2018.
\newblock URL \url{https://doi.org/10.1098/rsif.2017.0937}.

\bibitem[Fend et~al.(2003)Fend, Bovet, Yokoi, and Pfeifer]{fend2003active}
Miriam Fend, Simon Bovet, Hiroshi Yokoi, and Rolf Pfeifer.
\newblock An active artificial whisker array for texture discrimination.
\newblock In \emph{IEEE/RSJ International Conference on Intelligent Robots and
  Systems}, volume~2, pages 1044--1049. IEEE, 2003.
\newblock URL \url{https://ieeexplore.ieee.org/document/1248782}.

\bibitem[Finn et~al.(2017)Finn, Abbeel, and Levine]{finn2017model}
Chelsea Finn, Pieter Abbeel, and Sergey Levine.
\newblock Model-agnostic meta-learning for fast adaptation of deep networks.
\newblock In \emph{Proceedings of the 34th International Conference on Machine
  Learning-Volume 70}, pages 1126--1135. JMLR.org, 2017.
\newblock URL \url{http://proceedings.mlr.press/v70/finn17a/finn17a.pdf}.

\bibitem[French(1991)]{french1991using}
Robert~M French.
\newblock Using semi-distributed representations to overcome catastrophic
  forgetting in connectionist networks.
\newblock In \emph{Proceedings of the 13th annual cognitive science society
  conference}, pages 173--178. Erlbaum, 1991.
\newblock URL
  \url{https://www.aaai.org/Papers/Symposia/Spring/1993/SS-93-06/SS93-06-007.pdf}.

\bibitem[French(1999)]{french1999catastrophic}
Robert~M French.
\newblock Catastrophic forgetting in connectionist networks.
\newblock \emph{Trends in cognitive sciences}, 3\penalty0 (4):\penalty0
  128--135, 1999.
\newblock URL \url{https://doi.org/10.1016/S1364-6613(99)01294-2}.

\bibitem[Gidaris and Komodakis(2018)]{gidaris2018dynamic}
Spyros Gidaris and Nikos Komodakis.
\newblock Dynamic few-shot visual learning without forgetting.
\newblock In \emph{Proceedings of the IEEE Conference on Computer Vision and
  Pattern Recognition}, pages 4367--4375, 2018.
\newblock URL
  \url{http://openaccess.thecvf.com/content_cvpr_2018/papers/Gidaris_Dynamic_Few-Shot_Visual_CVPR_2018_paper.pdf}.

\bibitem[Goodfellow et~al.(2013)Goodfellow, Mirza, Xiao, Courville, and
  Bengio]{goodfellow2013empirical}
Ian~J Goodfellow, Mehdi Mirza, Da~Xiao, Aaron Courville, and Yoshua Bengio.
\newblock An empirical investigation of catastrophic forgetting in
  gradient-based neural networks.
\newblock \emph{arXiv preprint arXiv:1312.6211}, 2013.
\newblock URL \url{https://arxiv.org/abs/1312.6211}.

\bibitem[Hadka and Reed(2013)]{hadka12b}
David Hadka and Patrick Reed.
\newblock Borg: An auto-adaptive many-objective evolutionary computing
  framework.
\newblock \emph{Evolutionary Computation}, 21\penalty0 (2):\penalty0 231--259,
  2013.
\newblock URL \url{https://ieeexplore.ieee.org/document/6793867}.

\bibitem[Hauser et~al.(2011)Hauser, Ijspeert, F{\"u}chslin, Pfeifer, and
  Maass]{hauser2011towards}
Helmut Hauser, Auke~J Ijspeert, Rudolf~M F{\"u}chslin, Rolf Pfeifer, and
  Wolfgang Maass.
\newblock Towards a theoretical foundation for morphological computation with
  compliant bodies.
\newblock \emph{Biological cybernetics}, 105\penalty0 (5-6):\penalty0 355--370,
  2011.
\newblock URL \url{https://doi.org/10.1007/s00422-012-0471-0}.

\bibitem[He and Jaeger(2018)]{he2018overcoming}
Xu~He and Herbert Jaeger.
\newblock Overcoming catastrophic interference using conceptor-aided
  backpropagation.
\newblock In \emph{International Conference on Learning Representations}, 2018.
\newblock URL \url{https://openreview.net/pdf?id=B1al7jg0b}.

\bibitem[Kirkpatrick et~al.(2017)Kirkpatrick, Pascanu, Rabinowitz, Veness,
  Desjardins, Rusu, Milan, Quan, Ramalho, Grabska-Barwinska,
  et~al.]{kirkpatrick2017overcoming}
James Kirkpatrick, Razvan Pascanu, Neil Rabinowitz, Joel Veness, Guillaume
  Desjardins, Andrei~A Rusu, Kieran Milan, John Quan, Tiago Ramalho, Agnieszka
  Grabska-Barwinska, et~al.
\newblock Overcoming catastrophic forgetting in neural networks.
\newblock \emph{Proceedings of the National Academy of Sciences}, 114\penalty0
  (13):\penalty0 3521--3526, 2017.
\newblock URL \url{https://www.pnas.org/content/114/13/3521}.

\bibitem[Kramer et~al.(2011)Kramer, Majidi, and Wood]{kramer2011wearable}
Rebecca~K Kramer, Carmel Majidi, and Robert~J Wood.
\newblock Wearable tactile keypad with stretchable artificial skin.
\newblock In \emph{IEEE International Conference on Robotics and Automation},
  pages 1103--1107. IEEE, 2011.
\newblock URL \url{https://ieeexplore.ieee.org/document/5980082}.

\bibitem[Kriegman et~al.(2019)Kriegman, Walker, Shah, Levin, Kramer-Bottiglio,
  and Bongard]{kriegman2019automated}
Sam Kriegman, Stephanie Walker, Dylan Shah, Michael Levin, Rebecca
  Kramer-Bottiglio, and Josh Bongard.
\newblock Automated shapeshifting for function recovery in damaged robots.
\newblock In \emph{Proceedings of Robotics: Science and Systems}, 2019.
\newblock URL \url{http://www.roboticsproceedings.org/rss15/p28.pdf}.

\bibitem[Lichtensteiger and Eggenberger(1999)]{lichtensteiger1999evolving}
Lukas Lichtensteiger and Peter Eggenberger.
\newblock Evolving the morphology of a compound eye on a robot.
\newblock In \emph{Advanced Mobile Robots, 1999.(Eurobot'99) 1999 Third
  European Workshop on}, pages 127--134. IEEE, 1999.
\newblock URL \url{https://ieeexplore.ieee.org/document/827631}.

\bibitem[Lipson and Pollack(2000)]{lipson2000automatic}
Hod Lipson and Jordan~B Pollack.
\newblock Automatic design and manufacture of robotic lifeforms.
\newblock \emph{Nature}, 406\penalty0 (6799):\penalty0 974, 2000.
\newblock URL \url{https://www.nature.com/articles/35023115}.

\bibitem[Masse et~al.(2018)Masse, Grant, and Freedman]{masse2018alleviating}
Nicolas~Y Masse, Gregory~D Grant, and David~J Freedman.
\newblock Alleviating catastrophic forgetting using context-dependent gating
  and synaptic stabilization.
\newblock \emph{Proceedings of the National Academy of Sciences}, 115\penalty0
  (44):\penalty0 E10467--E10475, 2018.
\newblock URL \url{https://www.pnas.org/content/115/44/E10467}.

\bibitem[Narang et~al.(2018)Narang, Degirmenci, Vlassak, and
  Howe]{narang2018transforming}
Yashraj~S Narang, Alperen Degirmenci, Joost~J Vlassak, and Robert~D Howe.
\newblock Transforming the dynamic response of robotic structures and systems
  through laminar jamming.
\newblock \emph{IEEE Robotics and Automation Letters}, 3\penalty0 (2):\penalty0
  688--695, 2018.
\newblock URL \url{https://ieeexplore.ieee.org/document/8141952}.

\bibitem[Pervan and Murphey(2019)]{PERVAN2019197}
Ana Pervan and Todd Murphey.
\newblock Algorithmic materials: Embedding computation within material
  properties for autonomy.
\newblock In \emph{Robotic Systems and Autonomous Platforms}, pages 197 -- 221.
  Woodhead Publishing, 2019.
\newblock URL
  \url{http://www.sciencedirect.com/science/article/pii/B9780081022603000093}.

\bibitem[Powers et~al.(2018)Powers, Kriegman, and Bongard]{powers2018effects}
Joshua Powers, Sam Kriegman, and Josh Bongard.
\newblock The effects of morphology and fitness on catastrophic interference.
\newblock In \emph{Artificial Life Conference Proceedings}, pages 606--613,
  2018.
\newblock URL
  \url{https://www.mitpressjournals.org/doi/pdf/10.1162/isal_a_00111}.

\bibitem[Robins(1995)]{robins1995catastrophic}
Anthony Robins.
\newblock Catastrophic forgetting, rehearsal and pseudorehearsal.
\newblock \emph{Connection Science}, 7\penalty0 (2):\penalty0 123--146, 1995.
\newblock URL \url{https://doi.org/10.1080/09540099550039318}.

\bibitem[Schwarz et~al.(2018)Schwarz, Czarnecki, Luketina, Grabska-Barwinska,
  Teh, Pascanu, and Hadsell]{schwarz2018progress}
Jonathan Schwarz, Wojciech Czarnecki, Jelena Luketina, Agnieszka
  Grabska-Barwinska, Yee~Whye Teh, Razvan Pascanu, and Raia Hadsell.
\newblock Progress \& compress: A scalable framework for continual learning.
\newblock In \emph{Proceedings of the 35th International Conference on Machine
  Learning}, volume~80, pages 4528--4537. PMLR, 2018.
\newblock URL \url{http://proceedings.mlr.press/v80/schwarz18a/schwarz18a.pdf}.

\bibitem[Titsias et~al.(2019)Titsias, Schwarz, Matthews, Pascanu, and
  Teh]{titsias2019functional}
Michalis~K Titsias, Jonathan Schwarz, Alexander G de~G Matthews, Razvan
  Pascanu, and Yee~Whye Teh.
\newblock Functional regularisation for continual learning.
\newblock \emph{arXiv preprint arXiv:1901.11356}, 2019.
\newblock URL \url{https://arxiv.org/abs/1901.11356}.

\bibitem[Vaughan et~al.(2004)Vaughan, Di~Paolo, and
  Harvey]{vaughan2004evolution}
E~Vaughan, Ezequiel~A Di~Paolo, and I~Harvey.
\newblock The evolution of control and adaptation in a 3d powered passive
  dynamic walker.
\newblock In \emph{Proceedings of the Ninth International Conference on the
  Simulation and Synthesis of Living Systems, Artificial Life IX}, pages
  139--145, 2004.
\newblock URL
  \url{http://www.cs.uvm.edu/~jbongard/2015_CS206/2004_Vaughan.pdf}.

\end{thebibliography}
\end{document}